# A Method for Speeding Up Value Iteration in Partially Observable Markov Decision Processes

Nevin L. Zhang, Stephen S. Lee, and Weihong Zhang
Department of Computer Science, Hong Kong University of Science & Technology
{lzhang, sslee, wzhang}@cs.ust.hk

## Abstract

We present a technique for speeding up the convergence of value iteration for partially observable Markov decisions processes (POMDPs). The underlying idea is similar to that behind modified policy iteration for fully observable Markov decision processes (MDPs). The technique can be easily incorporated into any existing POMDP value iteration algorithms. Experiments have been conducted on several test problems with one POMDP value iteration algorithm called incremental pruning. We find that the technique can make incremental pruning run several orders of magnitude faster.

## 1 INTRODUCTION

POMDPs are a model for sequential decision making problems where effects of actions are nondeterministic and the state of the world is not known with certainty. They have attracted many researchers in Operations Research and AI because of their potential applications in a wide range of areas (Monahan 1982, Cassandra 1998b). However, there is still a significant gap between this potential and actual applications, primarily due to the lack of effective solution methods. For this reason, much recent effort has been devoted to finding efficient algorithms for POMDPs.

This paper is concerned with only exact algorithms. Most exact algorithms are value iteration algorithms. They begin with an initial value function and improve it iteratively until the Bellman residual falls below a predetermined threshold. See Cassandra (1998a) for excellent descriptions, analyses, and empirical comparisons of those algorithms.

There are also policy iteration algorithms for POMDPs. The first one is proposed by Sondik (1978). A simpler one is recently developed by Hansen (1998).

It is known that, in terms of number of iterations, policy iteration for MDP converges quadratically while value iteration converges linearly (e.g. Puterman 1990, page 369). Hansen has empirically shown that his policy iteration algorithm for POMDPs also converges much faster than one of the most efficient known value iteration algorithms, namely incremental pruning (Zhang and Liu 1997, Cassandra et al 1997).

Policy iteration for MDPs solves a system of linear equations at each iteration. The numbers of unknowns and equations in the system are the same as the size of the state space. Consequentially, it is computationally prohibitive to solve the system when the state space is large. Modified policy iteration (MPI) (Puterman 1990, page 371) alleviates the problem using a method that computes an approximate solution without actually solving the system. Numerical results reported in Puterman and Shin (1978) suggest that modified policy iteration is more efficient than either value iteration or policy iteration in practice.

Hansen (1998) points out that the idea of MPI can also be incorporated into his POMDP policy iteration algorithm and finds that such an exercise is not very helpful (Hansen 1999).

The paper describes another way to apply the MPI idea to POMDPs. Our method is based on the view that MPI is also a variant of value iteration (van Nunen 1976) [1]. Under this view, the basic idea is to "improve" the current value function for several steps using the current policy before feeding it to the next step of value iteration. Those improvement steps are less expensive than standard value iteration steps. Nonetheless, they do get the current value function closer to the optimal value function.

MPI for MDPs improves a value function at all states.

---

[1] As a matter of fact, it was first proposed as a variant of value iteration by van Nunen.

This cannot be done for POMDPs since there are infinite many belief states. Our method improves a value function at a finite number of selected belief states. A nice property of POMDPs is that when a value function is improved at one belief state, it is also improved in the neighborhood of that belief state.

We call our method *point-based improvement* for the lack of a better name. It is conceptually much simpler than Hensen's policy iteration algorithm. Nonetheless, it is as effective as, in some cases more effective than, Hansen's algorithm in reducing the number of iterations it takes to find a policy of desired quality and hence drastically speeds up incremental pruning.

## 2 POMDP AND VALUE ITERATION

We begin with a brief review of POMDPs and value iteration.

### 2.1 POMDPs

A *partially observable Markov decision process* (POMDP) is a sequential decision model for an agent who acts in a stochastic environment with only partial knowledge about the state of the environment. The set of possible states of the environment is referred to as the *state space* and is denoted by $\mathcal{S}$. At each point in time, the environment is in one of the possible states. The agent does not directly observe the state. Rather, it receives an observation about it. We denote the set of all possible observations by $\mathcal{Z}$. After receiving the observation, the agent chooses an action from a set $\mathcal{A}$ of possible actions and execute that action. Thereafter, the agent receives an immediate reward and the environment evolves stochastically into a next state.

Mathematically, a POMDP is specified by the three sets $\mathcal{S}$, $\mathcal{Z}$, and $\mathcal{A}$; a *reward function* $r(s,a)$; a *transition probability* $P(s'|s,a)$; and an *observation probability* $P(z|s',a)$. The reward function characterizes the dependency of the immediate reward on the current state $s$ and the current action $a$. The transition probability characterizes the dependency of the next state $s'$ on the current state $s$ and the current action $a$. The observation probability characterizes the dependency of the observation $z$ at the next time point on the next state $s'$ and the current action $a$.

### 2.2 Policies and value functions

Since the current observation does not fully reveal the identity of the current state, the agent needs to consider all previous observations and actions when choosing an action. Information about the current state contained in the current observation, previous observations, and previous actions can be summarized by a probability distribution over the state space. The probability distribution is sometimes called a *belief state* and denoted by $b$. For any possible state $s$, $b(s)$ is the probability that the current state is $s$. The set of all possible belief states is called the *belief space*. We denote it by $\mathcal{B}$.

A *policy* prescribes an action for each possible belief state. In other words, it is a mapping from $\mathcal{B}$ to $\mathcal{A}$. Associated with policy $\pi$ is its *value function* $V^\pi$. For each belief state $b$, $V^\pi(b)$ is the expected total discounted reward that the agent receives by following the policy starting from $b$, i.e.

$$V^\pi(b) = E_{\pi,b}[\sum_{t=0}^{\infty} \lambda^t r_t]$$

where $r_t$ is the reward received at time $t$ and $\lambda$ ($0 < \lambda < 1$) is the *discount factor*. It is known that there exists a policy $\pi^*$ such that $V^{\pi^*}(b) \geq V^\pi(b)$ for any other policy $\pi$ and any belief state $b$. Such a policy is called an *optimal policy*. The value function of an optimal policy is called the *optimal value function*. We denote it by $V^*$. For any positive number $\epsilon$, a policy $\pi$ is $\epsilon$-*optimal* if

$$V^\pi(b) + \epsilon \geq V^*(b) \quad \forall b \in \mathcal{B}.$$

### 2.3 Value iteration

To explain value iteration, we need to consider how belief state evolves over time. Let $b$ be the current belief state. The belief state at the next point in time depends not only on the current belief state, but also on the current action $a$ and the next observation $z$. We denote it by $b_z^a$. For any state $s'$, $b_z^a(s')$ is given by

$$b_z^a(s') = \frac{\sum_s P(s',z|s,a)b(s)}{P(z|b,a)}, \quad (1)$$

where $P(z,s'|s,a) = P(z|s',a)P(s'|s,a)$ and $P(z|b,a) = \sum_{s,s'} P(z,s'|s,a)b(s)$ is the renormalization constant. As the notation suggests, the constant can also be interpreted as the probability of observing $z$ after taking action $a$ in belief state $b$.

Define an operator $T$ that takes a value function $V$ and returns another value function $TV$ as follows:

$$TV(b) = max_a[r(b,a) + \lambda \sum_z P(z|b,a)V(b_z^a)] \forall b \in \mathcal{B} \quad (2)$$

where $r(b,a) = \sum_s r(s,a)b(s)$ is the expected immediate reward for taking action $a$ in belief state $b$. For a given value function $V$, a policy $\pi$ is said to be $V$-*improving* if

$$\pi(b) = arg\ max_a[r(b,a) + \lambda \sum_z P(z|b,a)V(b_z^a)] \quad (3)$$

for all belief states $b$.

Value iteration is an algorithm for finding $\epsilon$-optimal policies. It starts with an initial value function $V_0$ and iterates using the following formula:

$$V_n = TV_{n-1}.$$

It is known (e.g. Puterman 1990, Theorem 6.9) that $V_n$ converges to $V^*$ as $n$ goes to infinity. Value iteration terminates when the *Bellman residual* $max_b |V_n(b) - V_{n-1}(b)|$ falls below $\epsilon(1-\lambda)/2\lambda$. When it does, a $V_n$-improving policy is $\epsilon$-optimal.

Since there are infinite many possible belief states, value iteration cannot be carried explicitly. Fortunately, it can be carried out implicitly. Before explaining how, we first introduce several technical concepts and notations.

### 2.4 Technical and notational considerations

For convenience, we call functions over the state space *vectors*. We use lower case Greek letters $\alpha$ and $\beta$ to refer to vectors and script letters $\mathcal{V}$ and $\mathcal{U}$ to refer to sets of vectors. In contrast, the upper letters $V$ and $U$ always refer to value functions, i.e. functions over the belief space $\mathcal{B}$. Note that a belief state is a function over the state space and hence can be viewed as a vector.

A set $\mathcal{V}$ of vectors *induces* a value function as follows:

$$f(b) = max_{\alpha \in \mathcal{V}} \alpha.b \quad \forall b \in \mathcal{B},$$

where $\alpha.b$ is the inner product of $\alpha$ and $b$, i.e. $\alpha.b = \sum_s \alpha(s)b(s)$. For convenience, *we also use $\mathcal{V}(.)$ to denote the value function defined above: For any belief state $b$, $\mathcal{V}(b)$ stands for the quantity given at the right hand side of the above formula.*

A vector in a set is *extraneous* if its removal does not change the function that the set induces. It is *useful* otherwise. A set of vector is *parsimonious* if it contains no extraneous vectors.

### 2.5 Implicit value iteration

A value function $V$ is *represented* by a set of vectors if it equals the value function induced by the set. When a value function is representable by a finite set of vectors, there is a unique parsimonious set of vectors that represents the function.

VI:
1. $\mathcal{V}_0 \leftarrow \{0\}, n \leftarrow 0.$
2. do {
3. $\quad n \leftarrow n + 1.$
4. $\quad \mathcal{V}_n \leftarrow$ DP-UPDATE$(\mathcal{V}_{n-1})$.
5. } while $max_b|\mathcal{V}_n(b) - \mathcal{V}_{n-1}(b)| > \epsilon(1-\lambda)/2\lambda.$
6. return $\mathcal{V}_n$.

Figure 1: Value iteration for POMDPs.

Sondik (1971) has shown that if a value function $V$ is representable by a finite set of vectors, then so is the value function $TV$. The process of obtaining a parsimonious representation for $TV$ from a parsimonious representation of $V$ is usually referred to as *dynamic-programming update*. Let $\mathcal{V}$ be the parsimonious set that represents $V$. For convenience, we sometimes use $T\mathcal{V}$ to denote the parsimonious set of vectors that represents $TV$.

In practice, value iteration for POMDPs is implicitly carried in the way as shown in Figure 1. One begins with a value function $V_0$ that is representable by a finite set of vectors. In this paper, we assume the initial value function is 0. At each iteration, one performa dynamic-programming update on the parsimonious set $\mathcal{V}_{n-1}$ of vectors that represents the previous value function $V_{n-1}$ and obtains a parsimonious set of vectors $\mathcal{V}_n$ that represents the current value function $V_n$. One continues until the Bellman residual $max_b|\mathcal{V}_n(b) - \mathcal{V}_{n-1}(b)|$, which is determined by solving a sequence of linear programs, falls below a threshold.

## 3 PROPERTIES OF VALUE ITERATION

This paper presents a technique for speeding up the convergence of value iteration. The technique is designed for POMDPs with nonnegative rewards, i.e. POMDPs such that $r(s,a) \geq 0$ for all $s$ and $a$. In this section, we study the properties of value iteration in such POMDPs and show how a POMDP with negative rewards can be transformed into one with nonnegative rewards that is in some sense equivalent. Most proofs are omitted due to space limit.

We begin with a few definitions. In a POMDP, a value function $U$ *dominates* another $V$ if $U(b) \geq V(b)$ for all belief states $b$. It *strictly dominates* $V$ if it dominates $V$ and $U(b) > V(b)$ for at least one belief state $b$. A value function is *(strictly) suboptimal* if it is (strictly) dominated by the optimal value function.

A set of vectors is *(strictly) dominated* by a value function if its induced value function is. A set of vectors is *(strictly) suboptimal* if it is (strictly) dominated by the optimal value function.

A set of vectors is *(strictly) dominated* by another set of vectors if it is (strictly) dominated by the value function induced by the that set.

### 3.1 General properties of value iteration

**Lemma 1** *In any POMDP, if a set of vectors $\mathcal{V}$ is suboptimal, then so is $T\mathcal{V}$. Moreover, if $\mathcal{V}$ dominates another set of vectors $\mathcal{V}'$, then $T\mathcal{V}$ dominates $T\mathcal{V}'$.*

**Lemma 2** *In any POMDP, if a set of vectors $\mathcal{V}$ is strictly suboptimal, then there exist at least one belief state $b$ such that $T\mathcal{V}(b) > \mathcal{V}(b)$.*

### 3.2 Properties of value iteration in POMDPs with nonnegative rewards

Using Lemma 2, one can show

**Theorem 1** *Consider running VI on a POMDP with nonnegative rewards. Let $\mathcal{V}_{n-1}$ and $\mathcal{V}_n$ be respectively the sets of vectors produced at the n-1th and nth iteration. Then, $\mathcal{V}_{n-1}$ is strictly dominated by $\mathcal{V}_n$, which in turn is dominated by the optimal value function.*

Note that the theorem falls short of saying that, when the reward function is nonnegative, $T\mathcal{V}$ strictly dominates $\mathcal{V}$ for any suboptimal set of vectors $\mathcal{V}$. As a matter of fact, this is not always the case. As a counter example, assume $r(s_0, a) = 0$ for a certain state $s_0$ regardless of the action. Let $b_0$ be the belief state that is 1 at $s_0$ and 0 everywhere else. Further assume $V^*(b_0) > 0$ and let $\alpha_0$ be a vector such that $\alpha_0(s_0) = V^*(b_0)$ and $\alpha_0(s) = 0$ for any other states $s$. It is easy to see that if $\mathcal{V}$ consists of only $\alpha_0$, then $T\mathcal{V}(b_0) < \mathcal{V}(b_0)$.

Despite of the fact $T\mathcal{V}$ does not always strictly dominate $\mathcal{V}$, $T\mathcal{V}(b)$ is strictly larger than $\mathcal{V}(b)$ for beliefs $b$ in most parts of the belief space when the reward function is nonnegative.

### 3.3 POMDPs with negative rewards

A POMDP with negative rewards can always be transformed into one with nonnegative rewards by adding a large enough constant to the reward function. It is easy to see that an $\epsilon$-optimal policy for the transformed POMDP is also $\epsilon$-optimal for the original POMDP and vice versa. Moreover, the value function in the original POMDP of a policy equals that in the transformed POMDP minus $C/(1-\lambda)$, where $C$ is the constant added. In other words, the original POMDP is solved if the transformed POMDP is solved. Therefore, we can restrict to POMDPs with nonnegative rewards without losing generality.

VI1:
1. $\mathcal{V}_0 \leftarrow \{0\}, n \leftarrow 0$.
2. do {
3.     $n \leftarrow n + 1$;
4.     $\mathcal{U}_n \leftarrow$ DP-UPDATE($\mathcal{V}_{n-1}$);
5.     $\delta \leftarrow max_b |\mathcal{U}_n(b) - \mathcal{V}_{n-1}(b)|$;
6.     if $\delta > \epsilon(1-\lambda)/2\lambda$
7.        $\mathcal{V}_n \leftarrow$ improve($\mathcal{U}_n$);
8. } while $\delta > \epsilon(1-\lambda)/2\lambda$.
9. Return $\mathcal{V}_n$.

Figure 2: A new variant of value iteration.

Unless explicitly stated otherwise, all POMDPs considered from now on are with nonnegative rewards.

## 4 SPEEDING UP VALUE ITERATION

The section develops our technique for speeding up value iteration in POMDPs with nonnegative rewards. We begin with the basic idea.

### 4.1 Point-based improvement

Consider a suboptimal set of vectors $\mathcal{V}$. By *improving* $\mathcal{V}$, we mean to find another suboptimal set of vectors that strictly dominates $\mathcal{V}$. By *improving $\mathcal{V}$ at a belief state $b$*, we mean to find another suboptimal set of vectors $\mathcal{U}$ such that $\mathcal{U}(b) > \mathcal{V}(b)$.

Value iteration starts with the singleton set $\{0\}$, which is of course suboptimal, and improves the set iteratively using dynamic-programming update (Theorem 1). Dynamic-programming is quite expensive, especially when performed on large sets of vectors. To speed up value iteration, we devise a very inexpensive technique called *point-based improvement* for improving a set of vectors and use it multiple times in between dynamic-programming updates. This technique can be incorporated into value iteration as shown in Figure 2. Applications of the technique are encapsulated in the subroutine improve. The Bellman residual $\delta$ is used in improve to determine how many times the technique is to be used.

**Theorem 2** *If, for any suboptimal set of vectors $\mathcal{U}$, the output of improve($\mathcal{U}, \delta$) — another set of vectors — is suboptimal and dominates $\mathcal{U}$, then VI1 terminates in a finite number of steps.*

**Proof:** Let $\mathcal{V}_n$ and $\mathcal{V}'_n$ be respectively the sets of vectors produced at the $n$th iteration of VI1 and VI. From Lemma 1 and the condition imposed on improve, we conclude that $\mathcal{V}_n$ is suboptimal and dominates $\mathcal{V}'_n$.

Since $V'_n$ monotonically increases with $n$ (Theorem 1) and converges to $V^*$ as $n$ goes to infinity, $V_n$ must also converge to $V^*$. The theorem follows. □

## 4.2 Improving a set of vectors at one belief state

For the rest of this section, we let $V$ be a fixed suboptimal set of vectors and let $U=TV$. We develop a method for improving $U$.

To begin with, consider how $U$ might be improved at a particular belief state $b$. According to (2), there exists an action $a$ such that

$$U(b) = r(b,a) + \lambda \sum_z P(z|b,a)V(b_a^z). \quad (4)$$

For each observation $z$, let $\beta_z$ be a vector in $U$ that has maximum inner product with $b_a^z$. Define a new vector by

$$\beta(s) = r(s,a) + \lambda \sum_{z,s'} P(z,s'|s,a)\beta_z(s') \quad \forall s \in S. \quad (5)$$

We sometimes denote this vector by $\text{backup}(b,a,U)$.

**Theorem 3** *For the vector $\beta$ given by (5), we have*

$$b.\beta = r(b,a) + \lambda \sum_z P(z|b,a)U(b_a^z). \quad (6)$$

As pointed out at the end of Section 3.2, $U(b_a^z)$ is often larger than $V(b_a^z)$. A quick comparison of (4) and (6) leads us to conclude that $b.\beta$ is often larger than $U(b)$. When this is the case, we have found a set that improves $U$ at $b$, namely the singleton set $\{\beta\}$. The set is obviously suboptimal.

There is an obvious variation to the idea presented above. Instead of using the vector $\text{backup}(b,a,U)$ for the action that satisfies (4), we can consider the vectors $\text{backup}(b,a',U)$ for all possible actions $a'$ and choose the one whose inner product with $b$ is the maximum. This should, hopefully, improve $U$ at $b$ even further. We tried this variation and found that the costs are almost always greater than the benefits.

## 4.3 Improving a set of vectors at multiple belief states

It is straightforward to generalize the idea of the previous subsection from one belief state to multiple belief states. The question is what belief states to use. There are many possible answers. Our answer is motivated by the properties of dynamic-programming update.

For any vector $\alpha$ in $U$, define its *witness region* $R(\alpha,U)$ and *closed witness region* $\overline{R}(\alpha,U)$ w.r.t $U$ to be regions of the belief space $B$ respectively given by

$$R(\alpha,U) = \{b \in B | \alpha.b > \alpha'.b \quad \forall \alpha' \in U\setminus\{\alpha\}\},$$

$$\overline{R}(\alpha,U) = \{b \in B | \alpha.b \geq \alpha'.b \quad \forall \alpha' \in U\setminus\{\alpha\}\}.$$

During dynamic-programming update, each vector $\alpha$ in $U$ is associated with a belief state that is in the closed witness region of $\alpha$. We say that the belief state is the *anchoring point* provided for $\alpha$ by dynamic-programming update and denote it by $\text{point}(\alpha)$. The vector is also associated with an action, which we denote by $\text{action}(\alpha)$. It is the action prescribed for the belief state $\text{point}(\alpha)$ by a $V(.)$-improving policy. Because of those, equation (4) is true if $b$ is $\text{point}(\alpha)$ and $a$ is $\text{action}(\alpha)$.

We choose to improve $U$ on the anchoring points using

$$U_1 = \{\text{backup}(\text{point}(\alpha), \text{action}(\alpha), U) | \alpha \in U\}. \quad (7)$$

According to the discussions of the previous subsection, the value function $U_1(.)$ is often larger than $U(.)$ at the anchoring points. When a value function is larger than another one at one belief state, it is also larger in the neighborhood of the belief state. Therefore, the value function $U_1(.)$ is often larger than $U(.)$ in regions around the anchoring points. Our experience reveal that it is often larger in most parts of the belief space. The explanation is that the anchoring points scatter "evenly" over the belief space w.r.t $U$ in the sense that there is one in the closed witness region of each vector of $U$.

There is one optimization issue. Even that the inner product of the vector $\text{backup}(\text{point}(\alpha), \text{action}(\alpha), U)$ with the belief state $\text{point}(\alpha)$ is often larger than that of $\alpha$ with the belief state, it might be smaller sometimes. When this is the case, we use $\alpha$ instead of $\text{backup}(\text{point}(\alpha), \text{action}(\alpha), U)$ so that the value at the belief state is as large as possible.

## 4.4 Relation to modified policy iteration for MDPs

Point-based improvement is closely related to MPI for MDPs (Puterman 1990, page 371). It can be shown that, for each anchoring point $b$,

$$U_1(b) = r(b,\pi(b)) + \lambda \sum_z P(z|b,\pi(b))U(b_{\pi(b)}^z), \quad (8)$$

where $\pi$ is a $V(.)$-improving policy. This formula is very similar to formula (6.37) of Puterman (1990). MDP modified policy iteration uses the latter formula to "improve" the value of each possible state of the state space. We cannot apply the above formula to all possible belief states since there are infinite many of

```
improve(U):
1.  U_0 ← U, k ← 0.
2.  do {
3.    k ← k + 1, U_k ← ∅, W ← ∅.
4.    for each vector α in U_{k-1}
5.      α' ← backup(point(α),action(α),U_{k-1} ∪ W).
6.      if α.point(α) > α'.point(α)
7.        α' ← α.
8.      else W ← W ∪ {α'}.
9.      point(α') ← point(α),
10.     action(α') ← action(α);
11.     U_k ← U_k ∪ {α'}.
12. } while stop(U_k,U_{k-1}) = false.
13. return U_k ∪ U.
```

Figure 3: The improve subroutine.

them. So, we choose to use the formula to improve the values of a finite number of belief states, namely the anchoring points.

### 4.5 Repeated improvements

We now know how we might improve $U$ at the anchoring points. In hope to get as much improvement as possible, we want, of course, to apply the technique on $U_1$ and try to improve it further. This can easily be done. Observe that there is a one-to-one correspondence between vectors in $U$ and $U_1$: for each vector $\alpha$ in $U$, we have backup(point($\alpha$),action($\alpha$),$U$) in $U_1$. We associate the latter with the same belief state and action as the former. Then we can improve $U_1$ at the anchoring points the same way as we improve $U$. The process can of course be repeated for the resulting set of vectors and so on.

The above discussions lead to the routine shown in Figure 3. The routine improves the input vector set at the anchoring points iteratively. Improvement takes place at line 5. Lines 6 and 7 guarantee that the values of the anchoring point never decrease. The improved vector $\alpha'$ is added to $W$ at line 8 so that better improvements can be achieved for vectors yet to be processed. At lines 9 and 10, the belief state and action associated with a vector of the previous iteration are assigned to the corresponding vector at the current iteration.

The stopping criterion we use is

$max_{b:\text{an anchoring point}}[U_k(b) - U_{k-1}(b)] \leq \epsilon_1 \epsilon (1 - \lambda)/2\lambda$,

where $\epsilon_1$ is a positive number smaller than 1. In our experiments, $\epsilon_1$ is set at 0.. Compared with the stopping criterion of value iteration, the stopping criterion is stricter. The reason for this is that the improvement step is computationally cheap.

Finally, the union $U_k \cup U$ is returned instead of $U_k$ for the following reason. While $U_k(b)$ is no smaller than $U(b)$ at the anchoring points, it might be smaller at some other belief states. In other words, $U_k$ might not dominate $V$. If improve simply returns $U_k$, the conditions of Theorem 2 are not met. Consequently, the union $U_k \cup U$ is returned.

### 4.6 Pruning extraneous vectors

The union $U_k \cup U$ usually contains many extraneous vectors. They should be pruned to avoid unnecessary computations in the future. One way to doing so is to simply apply Lark's algorithm (White 1991).

Lark's algorithm solves a linear program for each input vector. It is expensive when there is a large number of vectors. We use a more efficient method. The motivation lies in two observations: First, most vectors in $U_k$ are not extraneous. Second, many vectors in $U$ are componentwise dominated by vectors in $U_k$ and hence are extraneous. The method is to replace line 13 with the following lines:

13. Prune from $U$ vectors that are componentwise dominated by vectors in $U_k$.
14. Prunes from $U$ vectors $\alpha$ such that $R(\alpha, U_k \cup U)$ is empty.
15. return $U_k \cup U$.

At line 14, a linear program is solved for each vector in $U$. Since no linear programs are solved for vectors in $U_k$ and the set $U$ usually becomes very small in cardinality after line 13, the method is much more efficient than simply applying Lark's algorithm to the union $U_k \cup U$.

### 4.7 Recursive calls to improve

Consider the set $U$ after line 14 of the algorithm segment given in the previous subsection. If it is not empty, then every vector $\alpha$ in the set is useful. This is determined by solving a linear program. In addition to determining the usefulness of $\alpha$, solving the linear program also produces a belief state $b$ that is in the closed witness region $\overline{R}(\alpha, U_k \cup U)$.

The facts that $\alpha$ is from the input set $U$ and that $b$ is in $\overline{R}(\alpha, U_k \cup U)$ imply that the value at $b$ has not been improved. To achieve as much improvement as possible, we improve the value by a recursive call to improve. To be more specific, we reset point($\alpha$) to $b$ at line 14 and replace line 15 with the following:

15. if $U \neq \emptyset$, return improve($U_k \cup U, \delta$).
16. else return $U_k$.

## 5   EMPIRICAL RESULTS

Experiments have been conducted to empirically determine the effectiveness of point-based improvement in speeding up value iteration and to compare it improvement with Hansen's policy iteration algorithm. Four problems were used in the experiments for both purposes. The problems are commonly referred to as Tiger, Network, Shuttle, and Aircraft ID in the literature and were downloaded from Cassandra's POMDP page [2]. Information about their sizes is summarized in the following table.

|  | $|\mathcal{S}|$ | $|\mathcal{Z}|$ | $|\mathcal{A}|$ |
|---|---|---|---|
| Tiger | 2 | 2 | 3 |
| Network | 7 | 2 | 4 |
| Shuttle | 8 | 2 | 3 |
| Aircraft ID | 12 | 5 | 6 |

### 5.1   Effectiveness of point-based improvement

The effectiveness of point-based improvement is determined by comparing VI and VI1. We borrowed an implementation of VI by Cassandra and VI1 was implemented on top of his program. Cassandra's program provides a number of algorithms for dynamic-programming update. For our experiments, we used a variation of incremental pruning called restricted region (Cassandra *et al* 1997). The discount factor was set at 0.95 and experiments were conducted on an UltraSparc II.

For the purpose of comparison, we collected information about the quality of the policies that VI and VI1 produce as a function of the times they take. The quality of a policy is measured by the distance between its value function to the optimal value function, i.e. the minimum $\epsilon$ such that the policy is $\epsilon$-optimal. The smaller the distance, the better the policy. Since we do not know the optimal value function, the distance cannot be exactly computed. We use an upper bound derived from the Bellman residual.

One experiment was conducted for each algorithm-problem combination. The experiment was terminated when either an 0.01-optimal policy was produced or the run time exceeded 24 hours, i.e. 86400 seconds, CPU time.

The data are summarized in the four charts in Figure 4. Note that both axes are in logarithmic scale. There is one chart for each problem. In each chart, there are two curves: one for VI and one for VI1. On each curve, there is data point for each iteration taken.

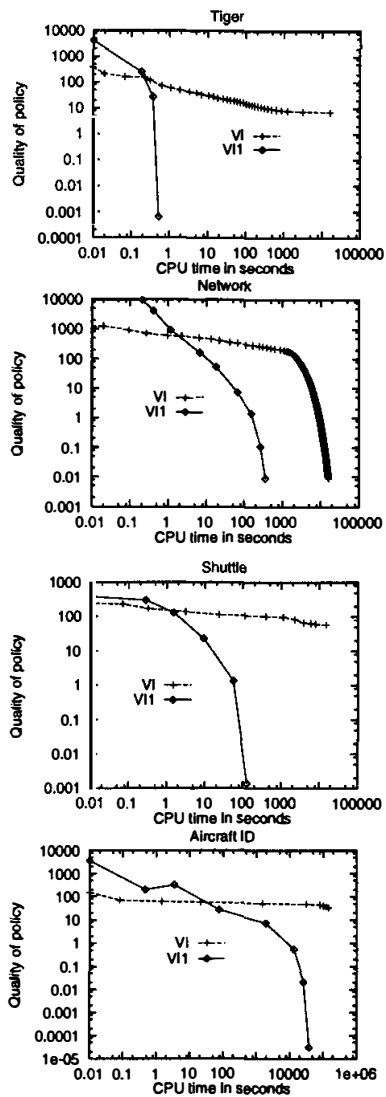

Figure 4: Empirical results. See text for explanations.

We see that VI1 was able to produce a 0.01-optimal policy for all four problems in a few iterations. On the other hand, VI took 215 iterations to produce a 0.01-optimal policy for Network. Within the time limit, VI completed only 35, 16, and 10 iterations respectively for Tiger, Shuttle, and Aircraft ID. Those suggest that the technique proposed in this paper is very effective in reducing the number of iterations that is required to produce good policies.

It is also clear that VI1 is much faster than VI. For Network, VI took about 17,000 seconds to produce a 0.01-optimal policy, while VI1 took only about 350 seconds. The speedup is about 50 times. VI was not able to produce "good" policies for Tiger, Shuttle, and Aircraft ID within the time limit, while VI1 produced 0.01-optimal or better policies for them in 0.53, 130, and 38,424 seconds respectively.

---
[2] http://www.cs.brown.edu/research/ai/pomdp/index.html

### 5.2 Comparisons with Hansen's policy iteration algorithm

In his implementation, Hansen used standard incremental pruning, instead of restricted region, for dynamic-programming update. Moreover, while the round-off threshold is set at $10^{-9}$ in Cassandra's program, Hansen set it at $10^{-6}$ probably because the routines for solving linear equations cannot handle precision beyond $10^{-6}$. For fairness of comparison, we implemented VI1 on top Hansen's code.

The following table shows the numbers of iterations and amounts of time VI1 and Hansen's algorithm took to find 0.01-optimal policies. We see that VI1 took fewer iterations than Hansen's algorithms on all problems. It took less on the first two problems and took roughly the same time on the last two problems.

|  | Iterations | | Time | |
| --- | --- | --- | --- | --- |
|  | VI1 | Hansen | VI1 | Hansen |
| Tiger | 4 | 14 | 0.51 | 3.3 |
| Network | 10 | 18 | 395 | 1122 |
| Shuttle | 6 | 9 | 65 | 73 |
| Aircraft ID | 8 | 9 | 72,377 | 66,964 |

## 6 CONCLUSIONS

We have developed a technique, namely point-based improvement, for speeding up the convergence of value iteration for POMDPs. The underlying idea is similar to that behind modified policy iteration for MDPs. The technique can easily be incorporated into any existing POMDP value iteration algorithms.

Experiments have been conducted on several test problems. We found that the technique is very effective in reducing the number of iterations that is required to obtain policies with desired quality. Because of this, it greatly speeds up value iteration. In our experiments, orders of magnitude speedups were observed.

**Acknowledgement**

Research is supported by Hong Kong Research Grants Council Grant HKUST6125/98E. The authors thank Cassandra and Hansen's for sharing with us their programs and the anonymous reviewers for useful comments.